\documentclass[final,3p]{elsarticle}

\usepackage{hyperref}
\usepackage{float}
\usepackage{verbatim} 
\usepackage{apalike}
\usepackage{graphicx}
\usepackage{amssymb}
\usepackage{amsthm}
\usepackage{amsmath}
\usepackage{caption}
\usepackage{subcaption}
\usepackage{booktabs}
\usepackage{url}
\usepackage{multirow}
\usepackage{hyperref}
\usepackage{lineno}   
\usepackage[lined,boxed,algoruled,longend,linesnumbered]{algorithm2e}
\usepackage[export]{adjustbox}
\usepackage{color}
\usepackage[table,xcdraw]{xcolor}
\usepackage{array}

\usepackage[normalem]{ulem}
\SetKw{Input}{input}
\SetKw{Output}{output}

\usepackage{color}

\usepackage{soul}

\biboptions{authoryear}

\begin{document}
\tolerance=999
\sloppy

\begin{frontmatter}


\title{Combining recurrent and residual learning for deforestation monitoring using multitemporal SAR images}

\author[1]{Carla Nascimento Neves}
\ead{cneves@lncc.br}
\author[2]{Raul Queiroz Feitosa}
\ead{raul@ele.puc-rio.br}
\author[2]{Mabel X. Ortega Adarme}
\ead{mortega@ele.puc-rio.br}
\author[1]{Gilson Antonio Giraldi}
\ead{gilson@lncc.br}

\address[1]{National Laboratory for Scientific Computing, Petropolis, RJ, Brazil} 
\address[2]{Pontifical Catholic University of Rio de Janeiro, 22451-900 Rio de Janeiro, Brazil}

\begin{abstract}

With its vast expanse, exceeding that of Western Europe by twice, the Amazon rainforest stands as the Earth's largest forest, holding immense importance in global climate regulation. Yet, deforestation detection from remote sensing data in this region poses a critical challenge, often hindered by the persistent cloud cover that obscures optical satellite data for much of the year. Addressing this need, this paper proposes three deep-learning models tailored for deforestation monitoring, utilizing SAR (Synthetic Aperture Radar) multitemporal data moved by its independence on atmospheric conditions. Specifically, the study proposes three novel recurrent fully convolutional network architectures—namely, RRCNN-1, RRCNN-2, and RRCNN-3—crafted to enhance the accuracy of deforestation detection. Additionally, this research explores replacing a bitemporal with multitemporal SAR sequences, motivated by the hypothesis that deforestation signs quickly fade in SAR images over time. A comprehensive assessment of the proposed approaches was conducted using a Sentinel-1 multitemporal sequence from a sample site in the Brazilian rainforest. The experimental analysis confirmed that analyzing a sequence of SAR images over an observation period can reveal deforestation spots undetectable in a pair of images. Notably, experimental results underscored the superiority of the multitemporal approach, yielding approximately a 5\% enhancement in F1-Score across all tested network architectures. Particularly the RRCNN-1 achieved the highest accuracy and also boasted half the processing time of its closest counterpart.

\end{abstract}


\end{frontmatter}


\section{Introduction}
\label{sec:introduction}

Remote sensing refers to acquiring information about an object from a remote location. This term is frequently employed to describe the imaging of the Earth's surface from an elevated perspective, such as via satellite \citep{parelius2023review}. Multitemporal remote sensing data can offer wide information for land change monitoring \citep{shi2020change, ban2016change}.

Change detection captures spatial differences in the state of an object by observing it at different times \citep{SINGH1989}. In the Remote Sensing context, its purpose is to monitor environmental changes by jointly processing a set of images of the same geographical area acquired at different dates, which is essential for the management of natural resources, the conservation of ecosystems and biodiversity as well as decision support for sustainable development \citep{asokan2019change}. 

Change detection using remote sensing imagery assumes a crucial function in numerous fields of applications, including disaster monitoring \citep{zheng2021building}, biodiversity study \citep{newbold2015global}, desertification \citep{dawelbait2012monitoring}, urbanization process \citep{han2017pre} and deforestation detection \citep{de2020change}, which is the focus of this research.

Among these applications, preserving the rainforests is critical in maintaining the health and stability of our planet's ecosystems. In particular, the Amazon rainforest, the Earth's largest forest, has suffered increasing deforestation rates in recent years, with Brazil being the country where the most significant statistics are concentrated \citep{giljum2022pantropical, amin2019neighborhood}. 

The Brazilian government tracks deforestation in the Amazon region through systematic satellite monitoring. For example, the Amazon Deforestation Monitoring Project (PRODES\footnote{http://www.obt.inpe.br/OBT/assuntos/programas/amazonia/prodes}) provides annual reports about deforestation in the Brazilian Legal Amazon (BLA) since 1988 \citep{valeriano2004}. One notable limitation of PRODES and several other deforestation monitoring systems is their dependency on optical data. Optical data are frequently hindered by cloud cover throughout most of the year in tropical regions \citep{doblas2020optimizing}


Various techniques have been addressed to create change maps for deforestation detection, including simple differencing \citep{stauffer1978landsat}, change vector analysis \citep{perbet2019near} and traditional machine learning techniques like Support Vector Machine \citep{01}, Principal Component Analysis \citep{sule2020application}, Random Forest \citep{02}, Maximum Likelihood \citep{03} and distance-based classifiers \citep{05}. 

Apart from these methods and their variations, numerous publications have incorporated deep learning in remote sensing change detection and demonstrated their superiority over conventional change detection methods \cite{bai2022deep}. 

Among the recent researches with deep learning solutions for change detection, Autoencoders, U-Net and its variants \citep{li2022densely, zheng2021clnet}, Recurrent Networks \citep{fang2023change, shi2022learning}, Generative Adversarial Networks \citep{zhao2020using}, and Transformer-based networks \citep{chen2021remote,wang2022network} are included.


A current research line \cite{panuju2020change} focuses on combining different techniques to improve change detection accuracy in remote sensing. Moreover, \cite{parelius2023review}  points out that the growing abundance of satellite imagery offers an opportunity to shift focus from the traditional bitemporal change detection methods used in previous studies to models that exploit longer time-series images. This approach allows for the incorporation of a richer set of data for change detection, a perspective that has been relatively underexplored until now, as emphasized in \cite{parelius2023review}.

The present study follows this trend and seeks to develop solutions for change detection with remote sensing data, specifically for deforestation monitoring, by employing multitemporal SAR data and combinations of different deep learning techniques. 
This paper proposes three recurrent fully convolutional deep networks for pixel-wise deforestation detection from SAR images. Furthermore, this study assesses the application of image sequences instead of bitemporal SAR image pairs for deforestation detection. This approach is inspired by the hypothesis that deforestation signs in tropical forests tend to rapidly diminish in SAR imagery due to the natural process of forest regeneration. Consequently, deforestation detection becomes increasingly challenging when bi-temporal image acquisition intervals are too widely apart. By harnessing a sequence of multiple images captured between such intervals, we substantially enhance the prospects of accurate and timely deforestation detection.


The main contributions of this work are:

\begin{itemize}
    \item Development of novel deep learning based solutions for automatic deforestation mapping and comparison with state-of-art methods.

    \item Addressing underexplored aspects in change detection literature, including the use of longer multitemporal data sequences. 

    \item Experimental analysis of the proposed methods using SAR data provided by Sentinel-1 from a sample site of the Amazon forest.
\end{itemize}

This paper is organized as follows. Section \ref{sec:relatedworks} discusses the recent research and also the gaps in change detection with deep learning. Section \ref{sec:theory} provides the theoretical background on the deep learning techniques that inspired the architectures developed in the present work and Section \ref{sec:proposal} introduces the proposed models. Section \ref{sec:methodology} presents the employed data, the experimental setup, and the architectures used for comparison. Section \ref{sec:results} shows and discusses the experimental results. Finally, Section \ref{sec:Concl} presents the main conclusions drawn from this study.

\section{Related Works}\label{sec:relatedworks}


With the continuous advancement of Deep Learning methods within computer vision, their application has been extended to the problem of change detection in Remote Sensing due to their ability to capture complex and hierarchical features present in the data \citep{parelius2023review}. Hence, this section will discuss recent works with deep learning solutions for change detection problems.


The U-Net encoder-decoder architecture and its variants are commonly employed for change detection tasks among the reviewed models. For instance, \cite{zheng2021clnet} introduced the Cross-Layer Convolutional Neural Network (CLNet), which is a modified U-Net with Cross-Layer Blocks (CLB) incorporated to integrate multi-scale features and multi-level contextual information by temporarily splitting the input into two parallel asymmetrical branches using different convolution strides and concatenating the feature maps from the two branches. Experiments were conducted on two building change detection datasets: the Learning Vision and Remote Sensing Laboratory building change detection (LEVIR-CD) and WHU-CD, reaching superior performance compared to several state-of-the-art methods. 

\cite{wang2022network} proposed the U-Net-like Visual Transformer for Change Detection (UVACD) for bitemporal image change detection. This network uses a CNN backbone for extracting semantic information followed by a visual transformer for feature enhancement that constructs change intensity tokens to complete the temporal information interaction and suppress irrelevant information weights to help obtain more distinguishable change features. The experiments were conducted on the WHU-CD and LEVIR-CD datasets. The authors reported a Precision of 94.58\%, a Recall of 91.17\%, an F1 score of 92.84\%, and an IoU of 86.64\% on the WHU-CD. UVACD outperformed some previous state of the art change detection methods in the experimental results.

U-Nets variants proposed in the literature take the form of a double-stream architecture. One example is the Densely Attentive Refinement Network (DARNet) introduced in \cite{li2022densely} to improve change detection on bitemporal very-high-resolution (VHR) remote sensing images. DARNet has a dense skip connections module between the encoder–decoder architecture, which combines features from various levels. A hybrid attention module is inserted at the skip connections level to combine temporal, spatial, and channel attention. Also, a recurrent refinement module is used to refine the predicted change in the decoding process. The experimental results on the season-varying change detection (SVCD) dataset, the Sun Yat-sen University change detection (SYSU-CD) dataset, and the LEVIR-CD dataset outperformed state-of-the-art models.

Some deep learning-based models proposed for change detection include Recurrent Neural Networks (RNN) due to their ability to handle related data sequences. The sequence usually consists of images from two different time points \citep{parelius2023review}.

\cite{fang2023change} proposed a fine-grained Multi-Functional Radar (MFR) model followed by a multi-head attention-based bi-directional Long Short Term Memory (LSTM) network to capture relationships between successive pulses. This process uses the temporal features to predict the probability of each pulse being a change point. The simulation results achieved better performance than the compared convolutional and recurrent networks.

\cite{shi2022learning} proposed a Multi-path Convolutional LSTM (MP-ConvLSTM) by combining LSTM and a CNN for change detection with bi-temporal hyperspectral images. A Siamese CNN was adopted to reduce the dimensionality of the images and extract preliminary features. A Convolutional LSTM was used to learn multi-level temporal dependencies among them. The MP-ConvLSTM was evaluated using four publicly available hyperspectral datasets acquired by the Earth Observing-1 (EO-1) Hyperion and obtained superior results than several state-of-the-art change detection algorithms, also exhibiting  better trade-off between complexity and accuracy in general. 

\cite{papadomanolaki2019detecting} presented a framework for urban change detection that combines a fully convolutional network (FCN) similar to U-Net for feature representation and a recurrent network for temporal modeling. The U-Net-based encoder–decoder architecture has a convolutional LSTM block added at all levels of the encoder. The authors evaluated the performance of this network using an ensemble cross-validation strategy on bi-temporal data from Onera Satellite Change Detection (OSCD) Sentinel-2 dataset (SAR data). The U-Net+LSTM model outperformed the regular U-Net.

Strategies using residual learning \citep{he2016deep} to facilitate gradient convergence are also applied with FCN approaches for change detection since such a combination helps obtain a more comprehensive range of information \citep{khelifi2020deep, shafique2022deep}.

\cite{basavaraju2022ucdnet} introduced UCDNet, the Urban Change Detection Network. This model is based on an encoder-decoder architecture, using a version of spatial pyramid pooling (SPP) blocks for extracting multiscale features and residual connections for introducing additional maps of feature differences between the streams at each level of the encoder to improve change localization, aiming to acquire better predictions while preserving the shape of changed areas. UCDNet uses a proposed loss function, a combination of weighted class categorical cross-entropy (WCCE) and modified Kappa loss. The authors evaluated the network on bi-temporal multispectral Sentinel-2 satellite images from Onera Satellite Change Detection (OSCD), obtaining better results than the models used for comparison..

The Multiscale Residual Siamese Network fusing Integrated Residual Attention (IRA-MRSNet), proposed by \cite{ling2022ira}, introduced multi-resolution blocks that combine convolutions with kernels of different sizes to extract deep semantic information and features at multiple scales. In addition, this network utilizes an attention unit connecting the encoder and the decoder. In experiments conducted on Seasonal Change Detection Dataset (CDD) the network outperformed the counterpart methods.

\cite{peng2019end} presented an improved U-Net++ design where change maps could be learned from scratch using available annotated datasets. The authors adopted the U-Net++ model with dense skip connections as the backbone for learning multiscale feature maps from several semantic layers. Residual blocks are employed in the convolution unit, aiming for better convergence. Deep supervision is implemented by using multiple side-output fusion (MSOF) to combine change maps from different semantic levels, generating a final change map. They used the weighted binary cross-entropy loss and the dice coefficient loss to mitigate the impact of class imbalance. The performance of the proposed CD method was verified on a VHR satellite image dataset and acheived a superior performance than the related methods.

Recent studies proposed change detection techniques by combining residual and recurrent learnings. For instance, \cite{khankeshizadeh2022fcd} presented FCD‐R2U‐Net, a forest change detection that includes a module for producing an enhanced forest fused difference image (EFFDI), to achieve a more efficient distinction of changes, followed by (Recurrent Residual U-Net) R2U-Net, applied to segment the EFFDI into the changed and unchanged areas. Experimental results were conducted on four bi-temporal images acquired by the Sentinel 2 and Landsat 8 satellite sensors. The qualitative and quantitative results demonstrated the effectiveness of the proposed EFFDI in reflecting the true forest changes from the background. Regarding the qualitative results, forest changes and their geometrical details were better preserved by FCD‐R2U‐Net, compared with U-Net, ResU-Net, and U-Net++. The proposed network also obtained superior results in the quantitative analysis. 

\cite{moustafa2021hyperspectral} proposed a change detection architecture named Attention Residual Recurrent U-Net (Att R2U-Net), inspired by R2U-Net and attention U-Net. This study supports the notion that deep neural networks can learn complex features and improve change detection performance when combined with hyperspectral data. Three hyperspectral change detection datasets with class imbalance and small regions of interest were employed to evaluate the performance of the proposed method for binary and multiclass change cases. The results were compared with U-Net, ResU-Net, R2U-Net, and Attention U-Net. Att R2U-Net outperformed the counterpart methods in almost all metrics and cases. 



As observed in the previously mentioned related works, in addition to developing new algorithms, combinations of available techniques are being considered to improve the accuracy of change detection in remote sensing. This is a research focus pointed out by \cite{panuju2020change}.




\section{Deep Learning Approaches Background}\label{sec:theory}

This section presents the deep learning approaches explored for constructing the change detection frameworks proposed in this study. The architectures are rooted in recurrent residual learning.

\subsection{Residual Networks}\label{sec:residual}

To enhance the training process of deep convolutional neural networks (CNNs), Residual Networks (ResNets) were conceived based on the observation that as neural networks grow deeper, they typically encounter elevated training errors, particularly when the network's depth becomes substantially large.

\begin{figure}[!ht]
\centering
\includegraphics[width=0.4\textwidth]{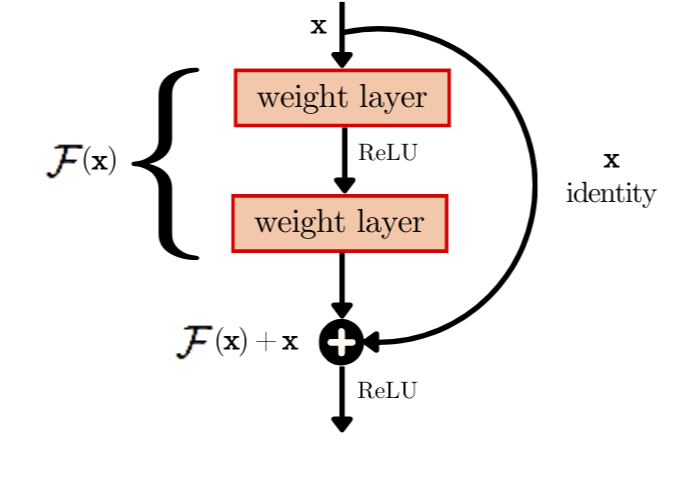}
\caption[Residual Learning]{Residual block (RB).}
\caption*{\footnotesize Source: Adapted from \cite{he2016deep}.}

\label{fig:resnet}
\end{figure}

\cite{he2016deep} introduced the so called residual blocks (see\ref{fig:resnet}  equipped with skip connections or shortcuts. In a residual block, the input to a layer is combined with the output of that layer, allowing the network to pass through information without significant alteration directly. As gradients backpropagate during training, they flow nearly unaltered through these skip connections, improving the convergence at the earlier layers. 

\cite{he2016deep} defines de building block as

\begin{equation}\label{eq:shortcut}
    \textrm{\textbf{y}} = \mathcal{F}(\textrm{\textbf{x}},\{\mathcal{W}_i\}) + \textrm{\textbf{x}} \textrm{ ,}
\end{equation}
where \textbf{x} and \textbf{y} are the input and output of the layers considered and the function $\mathcal{F}(\textrm{\textbf{x}},\{\mathcal{W}_i\})$  denotes the residual mapping with learnable weights assembled in $\mathcal{W}_i$. The example in Figure \ref{fig:resnet} has two layers, so $\mathcal{F} = W_2$ $\sigma(W_1\textrm{\textbf{x}})$, with $\sigma$ representing the rectified linear unit (ReLU) activation function \citep{nair2010rectified}. Biases are omitted for simplifying notations. It also seen in Figure \ref{fig:resnet} that the operation $\mathcal{F} + \textrm{\textbf{x}}$ is carried out by a shortcut connection and element-wise addition. These shortcut connections introduce no additional parameters and negligible computational complexity.

Inspired by the U-Net architecture and deep residual learning, \cite{zhang2018road} proposed the Deep Residual U-Net (ResU-Net). ResU-Net has three parts: The encoder consists of Residual Blocks (RB) and down-sampling operations that compress the input into compact representations. The bottleneck connects the encoder with the last part, the decoder, which comprises bilinear upsampling operations and recovers the representations to a pixel-wise categorization. In addition, skip connections link the first and last part so that information from the encoding layers is preserved and transmitted to the decoding layers.

According to \cite{zhang2018road}, the deep residual units make the deep network easy to train, and the skip connections within a residual unit and between the corresponding levels of the network will facilitate information propagation without degradation, making it possible to design a deep neural network with fewer parameters. 

As the present work approaches change detection, the decoder output feeds a softmax operator that delivers the posterior class probabilities for change or no change at each pixel location. The following figures show U-Net (Figure \ref{fig:unet}) and ResU-Net (Figure \ref{fig:resunet1}) variants used for deforestation detection in \citep{ortega2021comparison}.
The residual blocks (Figure \ref{fig:resunet2}) facilitate the flow of information through the network layers, allowing the capture of relevant features of the satellite image \citep{he2016deep}, which is crucial for change detection.


\begin{figure}[!htb]
     \centering
     \begin{subfigure}[b]{0.89\textwidth}
         \centering
         \includegraphics[width=0.75\textwidth]{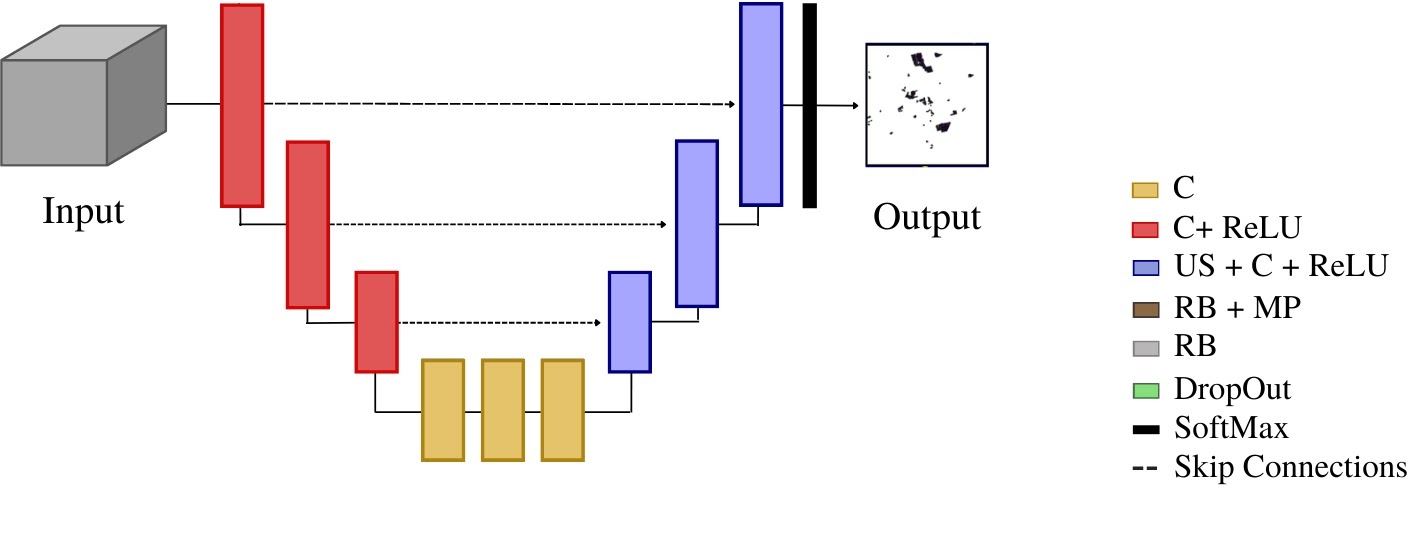}
         \caption[U-Net Architecture]{U-Net Architecture}
         \label{fig:unet}
     \end{subfigure}
     \hfill
     \\
     \begin{subfigure}[b]{0.25\textwidth}
         \centering
         \includegraphics[width=0.65\textwidth]{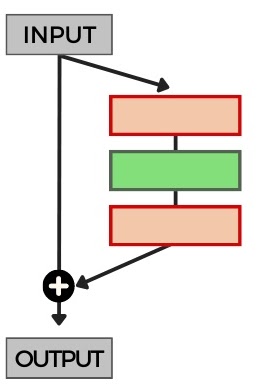}
         \caption[Residual Block]{Residual Block (RB)}
         \label{fig:resunet2}
     \end{subfigure}
     \hfill
     \begin{subfigure}[b]{0.63\textwidth}
         \centering
         \includegraphics[width=0.77\textwidth]{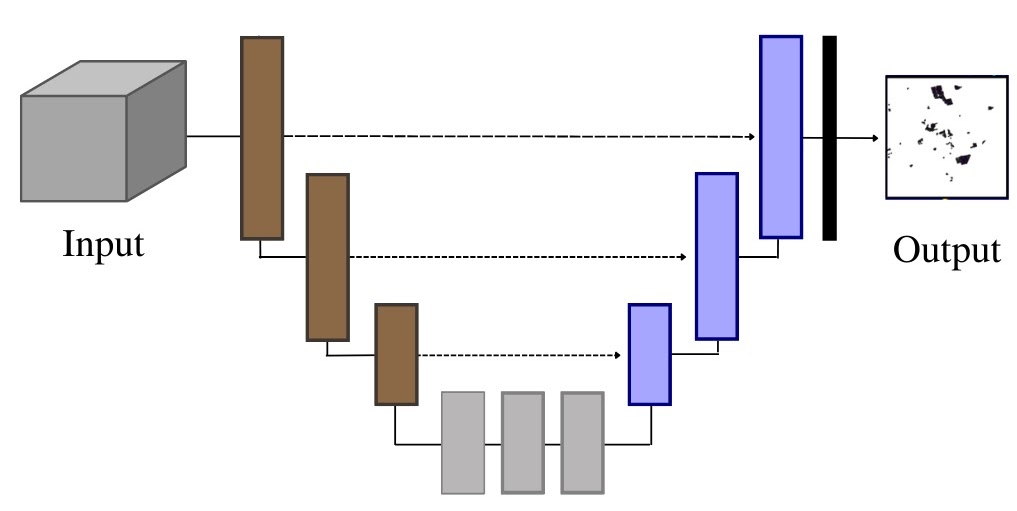}
         \caption[ResU-Net Architecture]{ResU-Net Architecture}
         \label{fig:resunet1}
     \end{subfigure}
        \caption[U-Net and ResU-Net Architectures]{U-Net and ResU-Net Architectures being used for a change detection scheme. Legend: C (Convolution), MP (Max-pooling), RB (Residual Block), US (Up-sampling).}
        \label{fig:resunet}
\end{figure}


\subsection{Recurrent Networks} \label{sec:convLSTM}

Recurrent Neural Networks (RNNs) are designed to process sequential data, updating their internal state at each time step $t$ while storing relevant information from prior steps \citep{rumelhart1986learning}. This enables RNNs to share weights across time steps, capturing temporal patterns \cite{goodfellowbook, graves2014generating}. However, the ability to retain information over long sequences is limited in traditional RNNs due to vanishing or exploding gradients during back-propagation \citep{calindeep}. The Long Short-term Memory (LSTM) \citep{hochreiter1997long} networks were developed to address such limitations of conventional RNNs. They incorporate gating mechanisms that selectively allow information to flow nearly unaltered through them, allowing the capture of long-range dependencies in sequential data.



\begin{figure}[!htb]
\centering
\includegraphics[width=0.5\textwidth]{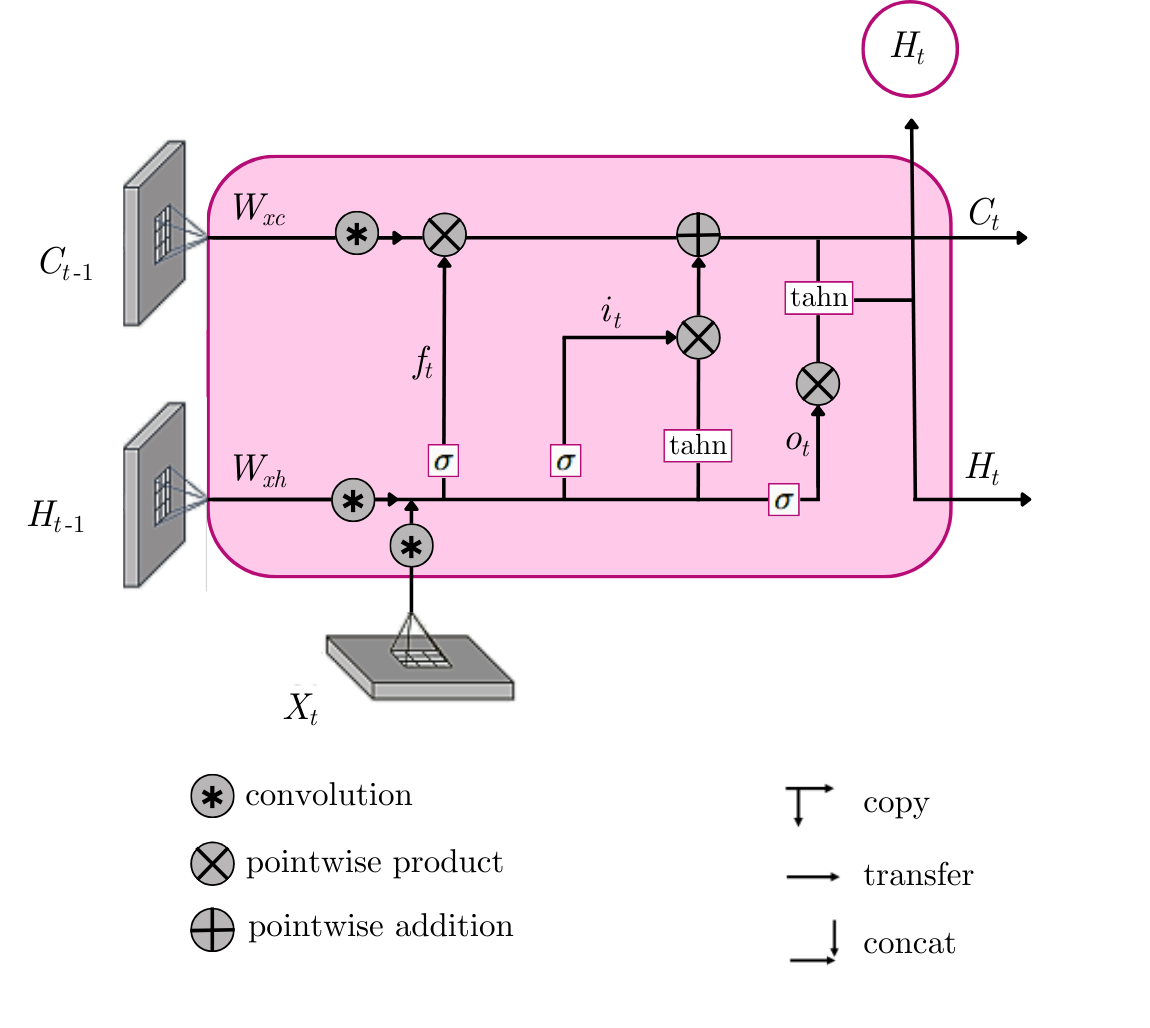}
\caption[ConvLSTM]{Inner structure of ConvLSTM}
\caption*{\footnotesize Source: Adapted from\cite{shi2022learning}}
\label{fig:convlstm}
\end{figure}

The present work employs the Convolutional LSTM Network (ConvLSTM). A ConvLSTM unit (See Figure \ref{fig:convlstm}) consists of inputs $X_t$, a cell state  $C_t$, a hidden state $H_t$ along with an input gate $i_t$, an output gate $o_t$ and a forget gate $f_t$ to control information flow. Due to the introduction of the convolutional structure, all the states, inputs, and intermediary outputs are three-dimensional tensors where the first two dimensions are spatial (rows and columns), and the last dimension learns feature representations \citep{shi2022learning}. The input $X_t$ and past states $C_{t-1}$, $H_{t-1}$ are employed to determine the future states $C_t$ and $H_t$.

The central equations are described in the sequence,  with '$*$' denoting the convolutional operator,  '$\circ$' the Hadamard product, $W$ the coefficient matrix, $\sigma$ the sigmoid function and $b$ the bias vector.

\begin{align}\label{eq:convLSTM}
    i_t &= \sigma(W_{xi} \ast X_t + W_{hi} \ast H_{t-1} + W_{ci} \circ C_{t-1} + b_i) \\
    f_t &= \sigma(W_{xf} \ast X_t + W_{hf} \ast H_{t-1} + W_{cf} \circ C_{t-1} + b_f) \\
    C_t &= f_t \circ C_{t-1} + i_t \circ \tanh(W_{xc} \ast X_t + W_{hc} \ast H_{t-1} + b_c) \\
    o_t &= \sigma(W_{xo} \ast X_t + W_{ho} \ast H_{t-1} + W_{co} \circ C_t + b_o) \\
    H_t &= o_t \circ \tanh(C_t)
\end{align}

According to \cite{shi2022learning}, the focus when using ConvLSTM for a change detection task is on capturing short-term temporal dependencies that accentuate bands capable of detecting changes while attenuating bands with less informative content. ConvLSTM recognizes and analyzes the multitemporal changes in image sequences  by capturing temporal dependencies and incorporating temporal features into the change detection process. Consequently, the hidden states $H_t$ of the ConvLSTM output can be extracted as representative features of changes. 

Considering this ability, convLSTM has been used in applications involving sequential images, such as detection of changes in hyperspectral images \citep{shi2022learning},  detection of urban changes \citep{papadomanolaki2019detecting} and deforestation detection \citep{masolele2021spatial}.

\subsection{Recurrent Residual Networks}\label{sec:res_rec}

According to \cite{yue2018residual}, residual learning and shortcut connections can effectively mitigate the exploding and vanishing gradient issues in long-term backpropagation. Bringing together the residual learning described in Section \ref{sec:residual} and the recurrent learning presented in Section \ref{sec:convLSTM}, Recurrent Residual Neural Networks have been proposed in the literature. Some examples include the Hybrid Residual LSTM (HRL) used for sequence classification \citep{wang2016recurrent}, R2U++ \citep{mubashar2022r2u++} and the Deep Recurrent U-Net (DRU-Net) \citep{kou2019microaneurysms} proposed for medical image segmentation.

\begin{figure}[!htb]
     \centering 
     \begin{subfigure}[b]{0.15\textwidth}
         \centering
         \includegraphics[scale=0.2]{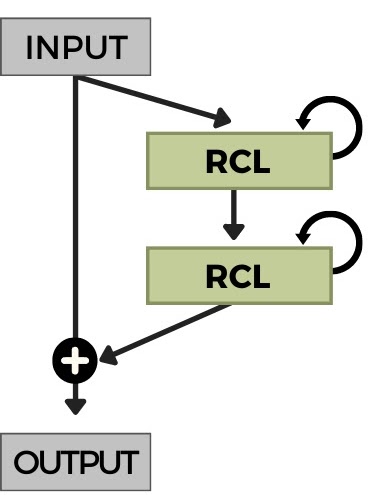}
         \caption[RRCU]{RRCU}
         \label{fig:rrcu}
     \end{subfigure}
     \hspace{1cm}
     \begin{subfigure}[b]{0.5\textwidth}
         \centering
         \includegraphics[scale = 0.2]{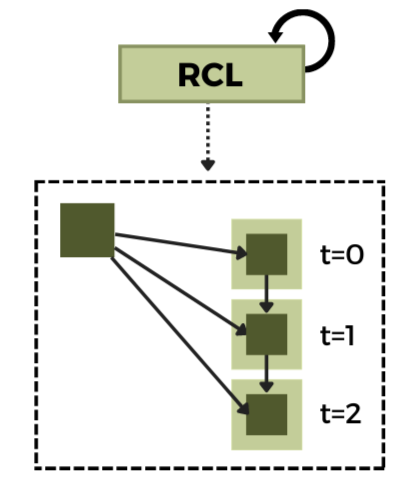}
         \caption{Unfolded Recurrent Convolutional Layer (RCL) with $t=2$}
         \label{fig:rcl}
     \end{subfigure}
     \hfill
        \caption[Recurrent Residual Convolutional Unit]{Recurrent Residual Convolutional Unit (RRCU)}
        \label{fig:rrcu_}
\end{figure}

A particular recurrent residual network called R2U-Net \citep{alom2019recurrent} takes the U-Net architecture and the residual blocks of ResU-Net (Figure \ref{fig:resunet}) as a starting point and includes the recurrent learning. This model combines recurrent convolutional operations with the residual blocks in Recurrent Residual Convolutional Units (RRCU - Figure \ref{fig:rrcu}) to replace the regular convolutional layers in the U-Net. Each RRCU has two Recurrent Convolutional Layers (RCL), and the input of the residual block is added to the output of the second RCL unit.  

The unfolded RCL (Figure \ref{fig:rcl}) for $t$ time steps is a feed-forward sub-network of depth $t+1$. The figure exemplifies an RCL with $t=2$, referring to the recurrent convolutional operation that includes one single convolution layer followed by two sub-sequential recurrent convolutional layers.

The following mathematical explanation of an RCL unit is adapted from \cite{liang2015recurrent}. For a unit located at $(i, j)$ on the $k$-th feature map in an RCL, the net input, $z_{ijk}(t)$ at a step $t$, is formulated as:

\begin{equation}\label{eq:rcl}
z_{ijk}(t) = (\textrm{\textbf{w}}^f_k)^T \textrm{\textbf{x}}^{f(i,j)}(t) + (\textrm{\textbf{w}}^r_k)^T \textrm{\textbf{x}}^{r(i,j)}(t - 1) + b_k \textrm{,}
\end{equation}
where \textbf{x}$^{f(i,j)}(t)$ and \textbf{x}$^{r(i,j)}(t-1)$ represents the feedforward and recurrent input, respectively. They correspond to the vectorized patches centered at $(i, j)$ of the feature maps in the current and previous layer. The terms \textbf{w}$^f_k$ and \textbf{w}$^r_k$ represent the vectorized feed-forward weights and recurrent weights, respectively, and $b_k$ is the bias. 
The first term in Eq. \ref{eq:rcl} is used in standard CNN and the recurrent connections induce the second term. The activity or state of this unit is a function of $z_{ijk}(t)$, where $\sigma$ is the ReLU:

\begin{equation}\label{eq:f(x,w)}
    \sigma(z_{ijk}(t)) = \textrm{max}(z_{ijk}(t), 0) 
\end{equation}

The RRCU proposed by \cite{alom2019recurrent} uses the RCL in a Residual Block, as shown in Figure \ref{fig:rrcu}.  Considering $x_l$ as the input in the $l^{th}$ layer of an RRCU, the output of this unit, $x_{l+1}$, can be calculated as:

\begin{equation}
x_{l+1} = x_l + F(x_l, w_l)\textrm{,}
\end{equation}
where $F(x_l, w_l)$ is the output of the last RCL, expressed as

\begin{equation}
    F(x_l, w_l) = \sigma(z^l_{ijk}(t)) = \textrm{max}(z^l_{ijk}(t), 0) \textrm{.}
\end{equation}

\begin{figure}[!ht]
\centering
\includegraphics[width=0.55\textwidth]{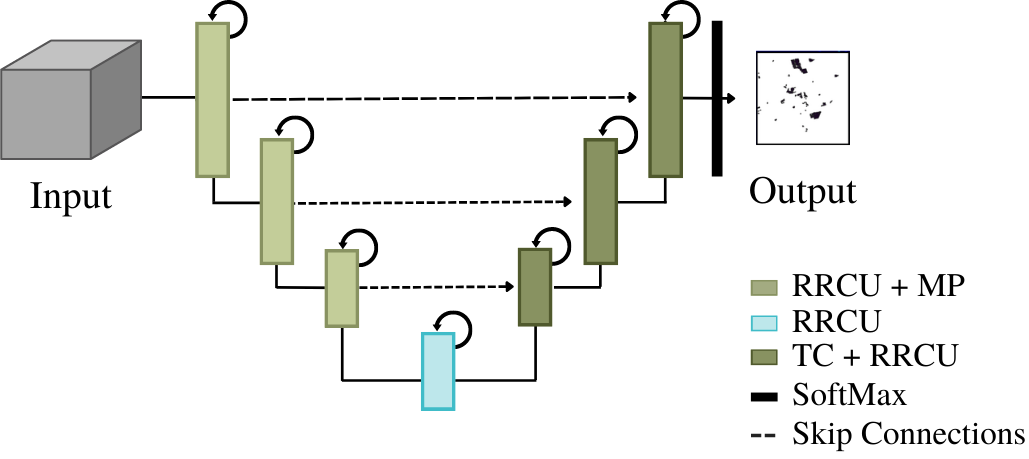}
\caption[R2U-Net Architecture]{R2U-Net Architecture for change detection. Legend: MP (Max-pooling), TC (Transpose Convolution), RRCU (Recurrent Residual Convolutional Unit)}
\label{fig:r2unet}
\end{figure}

Figure \ref{fig:r2unet} shows a R2U-Net architecture with three encoding and decoding levels and a bottleneck between these stages, employing RRCUs.

Recurrent residual architectures derived from  R2U-Net, including FCD‐R2U‐Net \citep{khankeshizadeh2022fcd} and Att R2U-Net \citep{moustafa2021hyperspectral} have been used for change detection tasks. \cite{alom2019recurrent} and  \cite{kou2019microaneurysms} state that the inclusion of RCLs in residual units further enhances the ability to handle deeper architectures. Also, the process of collecting and combining information from different time-steps in a recurrent neural network allows the model to capture dependencies and patterns over longer sequences of data. This feature accumulation process leads to more comprehensive feature representations and helps the model extract very low-level features from the data, which are crucial for segmentation tasks across various modalities.

\section{Proposed Architectures} \label{sec:proposal}

\begin{figure}[!ht]
\centering
\includegraphics[width=0.55\textwidth]{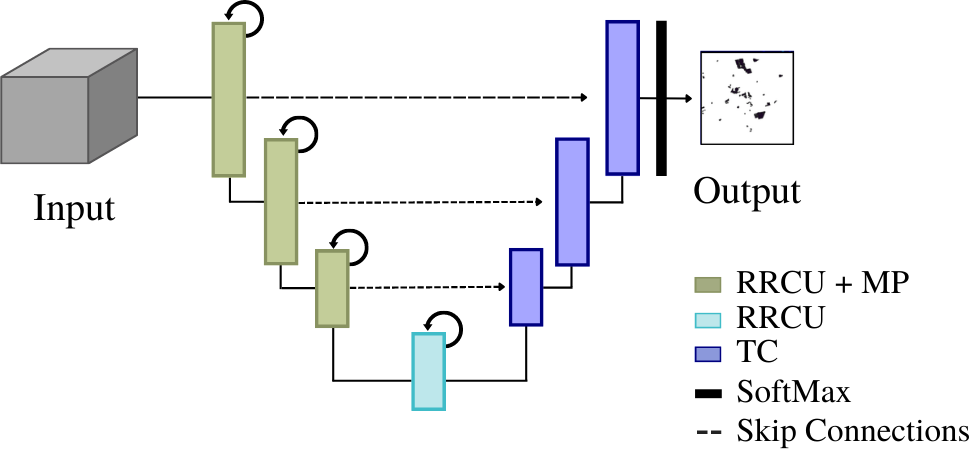}
\caption[RRCNN-1 Architecture]{Modified R2U-Net Architecture: RRCNN-1}
\label{fig:RRCNN1}
\end{figure}

This section presents the deep learning solutions for deforestation monitoring proposed in this work, drawing on the principles of recurrent residual learning discussed in Section \ref{sec:res_rec}. In this way, in addition to the ability to learn in multiple layers and the mitigation of the problem of vanishing gradients provided by the residual blocks, there will be benefits from the temporal dependency modeling provided by the recurrent layers. The first proposal, illustrated in Figure \ref{fig:r2unet},  consists of an adaptation of R2U-Net, using the RRCUs in the encoder and in the bottleneck and the replacement of these units in the decoder by transposed convolutions (TC), as shown in Figure \ref{fig:RRCNN1}. The result is a network with fewer parameters.

\begin{figure}[H]
     \centering
     \begin{subfigure}[b]{0.25\textwidth}
         \centering
         \includegraphics[scale = 0.2]{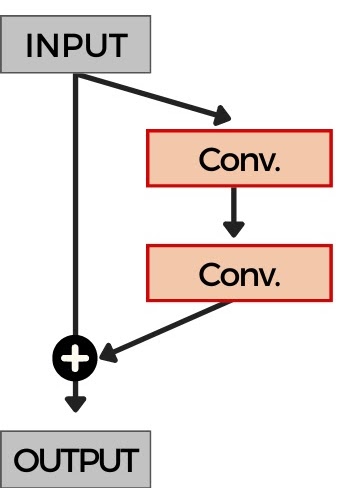}
         \caption{Typical Residual Block}
         \label{fig:resblock}
     \end{subfigure}
     \hfill
     \begin{subfigure}[b]{0.25\textwidth}
         \centering
         \includegraphics[scale = 0.2]{fig/rrcu.jpg}
         \caption{RRCU}
         \label{fig:rrcu2}
     \end{subfigure}
     \hfill
     \begin{subfigure}[b]{0.25\textwidth}
         \centering
         \includegraphics[scale = 0.2]{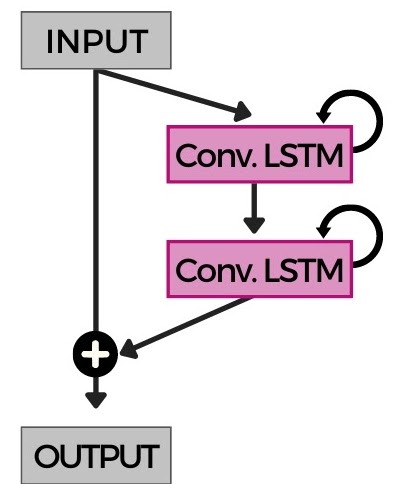}
         \caption{RCLSTM (proposed)}
         \label{fig:RCLSTM}
     \end{subfigure}
        \caption{Different variants of residual and recurrent residual convolutional units.}
        \label{fig:rrunits}
\end{figure}

The second proposal relies on a new recurrent residual block, the Residual Convolutional LSTM block (RCLSTM), that consists of convolutional LSTM blocks with a ReLU activation instead of the RCL used in R2U-Net's RRCU. 
Figure \ref{fig:rrunits} highlights the differences between the RCLSTM block proposed here and a typical residual block (Figure \ref{fig:resblock}) and an RRCU block (Figure \ref{fig:rrcu2}), already described in Section \ref{sec:res_rec}.

\begin{figure}[!ht]
\centering
\includegraphics[width=0.6\textwidth]{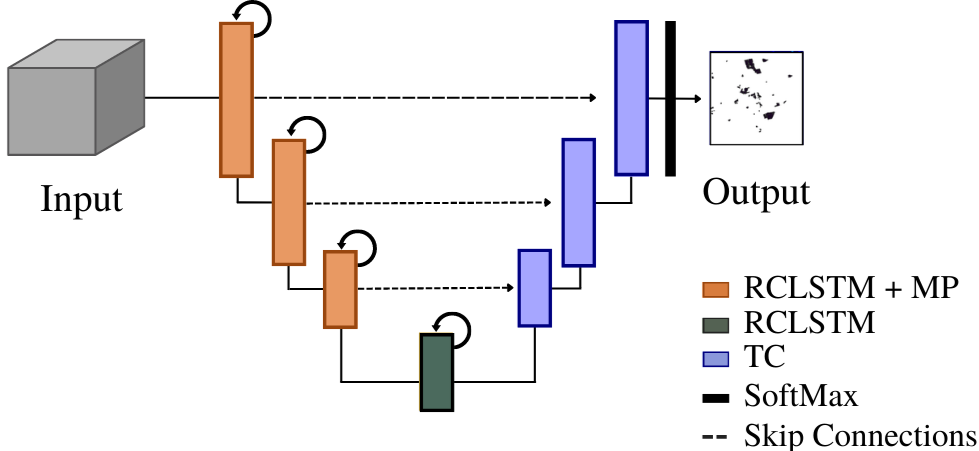}
\caption[RRCNN-2 Architecture]{RRCNN-2 Architecture.}
\label{fig:RRCNN2}
\end{figure}

The second proposal derives from the RRCNN-1, replacing the RRCUs by the RCLSTM block in the encoder and bottleneck. Figure \ref{fig:RRCNN2} depicts the proposed recurrent residual network hereafter called RRCNN-2.

\begin{figure}[!ht]
\centering
\includegraphics[width=0.6\textwidth]{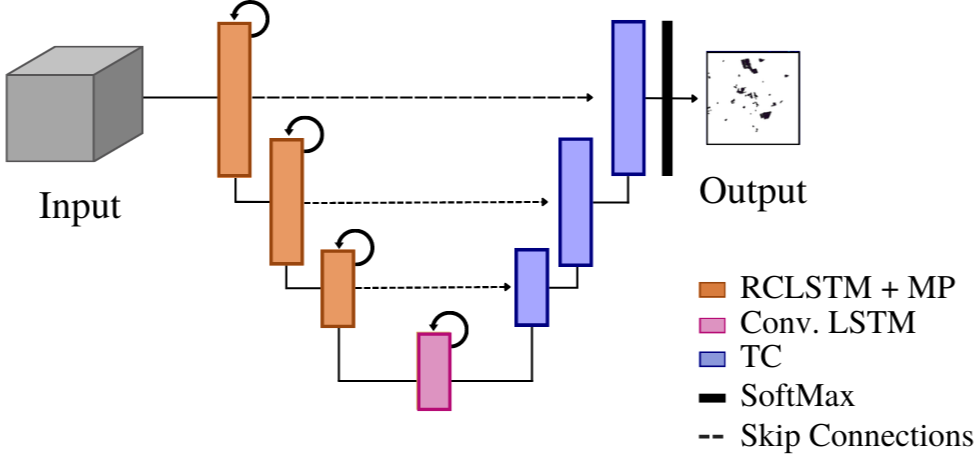}
\caption[RRCNN-3 Architecture]{RRCNN-3 Architecture.}
\label{fig:RRCNN3}
\end{figure}

The third proposed architecture proposed in this work, named  RRCNN-3, is a kind of variant of the prior architecture resulting from replacing the RCLSTM block in the bottleneck with a singular convolutional LSTM block, as depicted in Figure \ref{fig:RRCNN3}.

\section{Experimental Analysis}\label{sec:methodology}

This section presents the datasets and the experimental setup employed in this study. The deep learning methods presented in Section \ref{sec:theory} were used for comparison purposes with the architectures proposed in Section \ref{sec:proposal}.

\subsection{Dataset}\label{sec:dataset}

\begin{figure}[!htb]
\begin{center}
\includegraphics[scale=0.7]{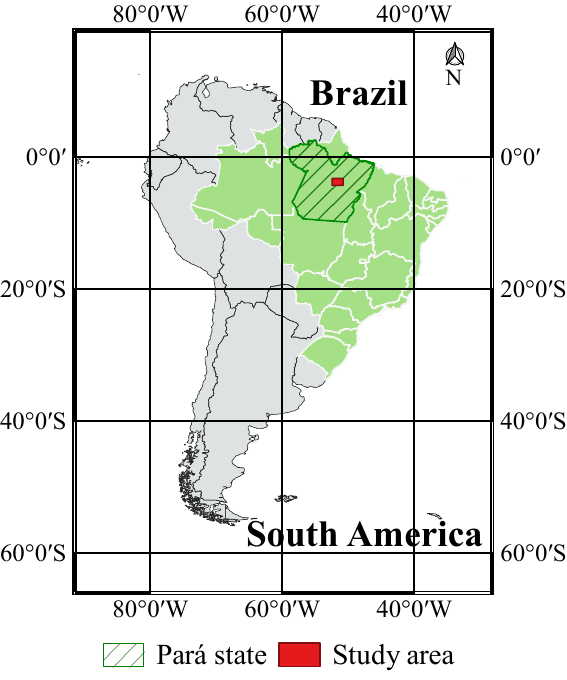}
\caption{Geographical location of the study site in the Pará state, Brazil.}
\label{fig:study_areas}
\end{center}
\end{figure}

This study employed Sentinel-1 data from a site in the Brazilian Legal Amazon in the Pará state that extends over $115 \times 186$ Km². The site is characterized by mixed land cover, mainly dense evergreen forests and pastures (see Figure \ref{fig:study_areas}).

\begin{figure}[!t]
\centering
\includegraphics[width=1\textwidth]{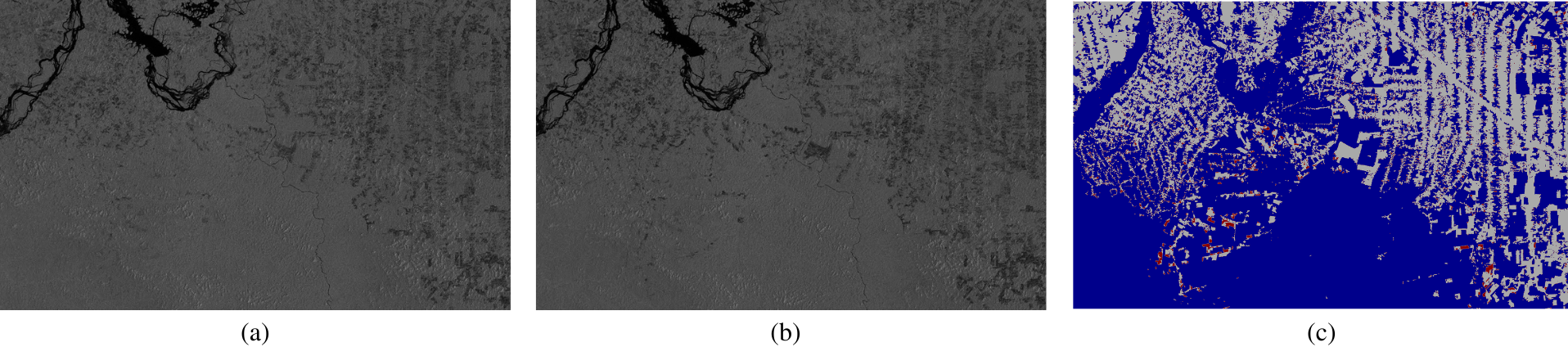}
\caption[SAR images and Ground Truth]{The SAR images: (a) Initial date;  (b) Final date; (c) Ground truth of the deforestation that occurred in the period. Legend - {\color[HTML]{656565} gray}: past deforestation (1988-2018); {\color[HTML]{9A0000} red}: deforestation (2019-2020); {\color[HTML]{00009B} blue}: no-deforestation.}
\label{fig:parachange}
\end{figure}

To build the input data for the deep learning models described in Section \ref{sec:theory} and Section \ref{sec:proposal}, seven images from this same geographical site with resolution of $9327 \times  5767 \times 2$ (width $\times$ height $\times$ polarizations - VV and VH) were captured with a periodicity of approximately two months, starting in August 2019 (Figure \ref{fig:parachange}(a)) and ending in August 2020 (Figure \ref{fig:parachange}(b)).

The reference map of the deforestation that occurred in this period is available on the INPE website\footnote{http://terrabrasilis.dpi.inpe.br/} (Figure \ref{fig:parachange}(c)). It is worth mentioning that this dataset is highly unbalanced, with only 1.06\% of the pixels belonging to the deforestation class, 34.04\% corresponding to past-deforestation class, and 64.9 \% to the no-deforestation class.

\begin{figure}[!ht]
\centering
\includegraphics[width=0.7\textwidth]{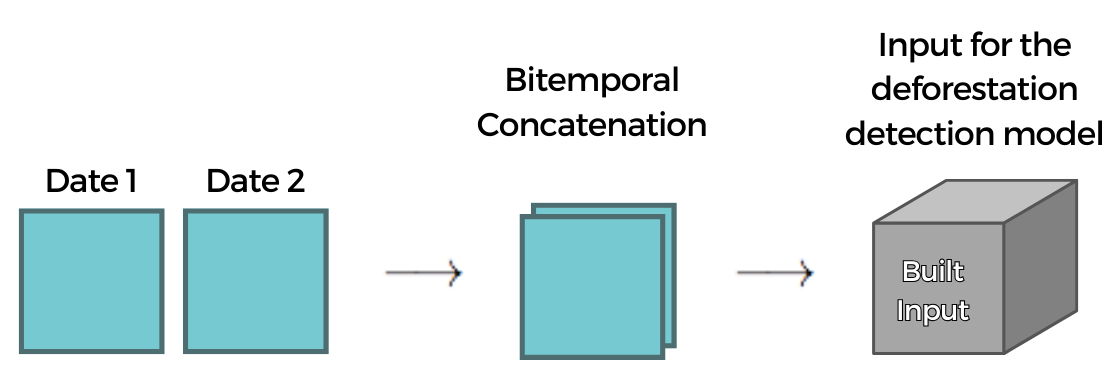}
\caption[Bitemporal Concatenation]{Model Input construction employed for the current experiments. Bitemporal example for data acquired from the same sensor.}
\label{fig:multitemporal_fusion}
\end{figure}

The input consists of a tensor $\mathbf{I} \in \mathbf{R}^{H\times W \times 2D}$ resulting from stacking $D$ multitemporal SAR images with 2 polarizations along the third dimension. In our experiments, we explored two scenarios: $D=2$ to represent the conventional bitemporal set and $D=7$ to represent an extended multitemporal sequence. 
Figure \ref{fig:multitemporal_fusion} shows this process for the bitemporal case.

Each input tensor was splited into 60 tiles of $932 \times 961$ pixels. A cross-validation strategy with six folds was adopted during the training. Each tile was part of the test set only once. Then, the final prediction was a mosaic of all test tiles covering the whole image.


During training and validation the network receives as input tensor patches of $128 \times 128$ pixels cropped from the training tiles, with maximum overlap of 70\% allowed. 

\subsection{Networks configuration and hyuperparameters}\label{sec:preparation} 

The experimental analysis reported in the next section compares the results obtained with the approaches discussed in Section \ref{sec:theory}, which serve as baselines, with the models proposed in Section \ref{sec:proposal}. Table \ref{tab:architectures} shows the architectures evaluated in our experiments.

\begin{table}[!ht]
\centering
\caption[Networks Architectures]{Networks Architectures}
\label{tab:architectures}
\setlength\extrarowheight{3pt}
\resizebox{\linewidth}{!}{
\begin{tabular}{l|c|c|c|r}
\hline
Architectures & Encoder             & Bottleneck          & Decoder           & Output                      \\ \hline
              & MP(C(3×3,32))       & C(3×3,128)          & US(C(3×3,128))    &                             \\
U-Net          & MP(C(3×3,64))       & C(3×3,128)          & US(C(3×3,64))     & Softmax (C(1×1, \#Classes)) \\
              & MP(C(3×3,128))      & C(3×3,128)          & US(C(3×3,32))     &                             \\ \hline
              & MP(RB(3×3,32))      & RB(3×3,128)         & US(C(3×3,128))   &                             \\
ResU-Net       & MP(RB(3×3,64))      & RB(3×3,128)         & US(C(3×3,64))    & Softmax (C(1×1, \#Classes)) \\
              & MP(RB(3×3,128))     & RB(3×3,128)         & US(C(3×3,32))    &                             \\ \hline
              & MP(RRCU(3×3,32))    &                     & US(RRCU(3×3,128)) &                             \\
R2U-Net        & MP(RRCU(3×3,64))    & RRCU(3×3,128)       & US(RRCU(3×3,64))  & Softmax (C(1×1, \#Classes)) \\
              & MP(RRCU(3×3,128))   &                     & US(RRCU(3×3,32))  &                             \\ \hline
              & MP(RRCU(3×3,32))    &                     & TC(3×3,128)       &                             \\
RRCNN-1       & MP(RRCU(3×3,64))    & RRCU(3×3,128)       & TC(3×3,64)        & Softmax (C(1×1, \#Classes)) \\
              & MP(RRCU(3×3,128))   &                     & TC(3×3,32)        &                             \\ \hline
              & MP(RCLSTM(3×3,32))  &                     & TC(3×3,128)       &                             \\
RRCNN-2       & MP(RCLSTM(3×3,64))  & RCLSTM(3×3,128)     & TC(3×3,64)        & Softmax (C(1×1, \#Classes)) \\
              & MP(RCLSTM(3×3,128)) &                     & TC(3×3,32)        &                             \\ \hline
              & MP(RCLSTM(3×3,32))  &                     & TC(3×3,128)       &                             \\
RRCNN-3       & MP(RCLSTM(3×3,64))  & Conv. LSTM(3×3,128) & TC(3×3,64)        & Softmax (C(1×1, \#Classes)) \\
              & MP(RCLSTM(3×3,128)) &                     & TC(3×3,32)        &                             \\ \hline
\end{tabular}
}
\caption*{\footnotesize The parametrization is (Kernel Height x Kernel Width, Number of filters). Symbols: C (Convolution), MP (Max-pooling), RB (Residual Block), US (Up-sampling), TC (Transpose Convolution), RRCU (Recurrent Residual Convolutional Unit), RCLSTM (Residual Convolutional LSTM block). }
\end{table}

The employed parameter setup follows: batch size equal to $32$, Adam optimizer with learning rate equal to $1e^{-3}$, and $\beta$ equal to $0.9$, and, to avoid over-fitting, an early stopping strategy with patience equal to $10$. 

Considering that the dataset is highly unbalanced, the emplowed loss function was the weighted categorical cross entropy, given by \cite{ho2019real}:

\begin{equation}
\textrm{Loss}_{wcce}  =    -\frac{1}{M} \sum^K_{k=1}\sum^M_{m=1} w_k \cdot y_m^k \cdot \textrm{log}(\hat{y}_m^k)\textrm{,}
\end{equation}
where $M$ is the number of training pixels, $K$ is the number of classes, $w_k$ is the weight for class $k$, $y_m^k$ is the target label for training example $m$ for class $k$, $x_m$ is the input for training example $m$ and $\hat{y}_m^k$ refers to the predicted probability for training example $m$ for class $k$.

In the present case, adopted the weights  $0.2$  for class no-deforestation and  $0.8$ for class deforestation. Following the PRODES methodology, the past deforestation class was ignored during training, validation, and testing. Only patches having at least 2\% of pixels of the deforestation class were used for training. In addition, a data augmentation procedure was applied for training and validation; these operations included rotation (multiples of 90º) and flipping (horizontal, vertical) transformations. The threshold to separate the deforestation and no-deforestation classes was 50\%.

\subsection{Evaluation metrics}\label{sec:metrics}

The generated deforestation maps classify each pixel into categories of deforestation and no-deforestation. Designating deforestation as a “positive” and no-deforestation as “negative”, there are four possible outcomes: true positive (TP) being a correctly identified deforestation/positive, true negative (TN) being a correctly identified no-deforestation, false positive (FP) being an unchanged pixel labeled as deforestation and false negative (FN) a changed pixel labeled as no-deforestation.

Among the several performance metrics that have been used to evaluate the results of a deforestation detection process, three of the most common are Precision, Recall, and F1-Score, as can be seen in the studies mentioned in Section\ref{sec:relatedworks} and also in review articles about change detection with remote sensing data \citep{parelius2023review, shafique2022deep}. 

The Precision metric denotes the ratio between the number of correctly classified deforestation pixels and the total number of pixels identified as deforestation.

\begin{equation}
    \textrm{Precision} = \frac{TP}{TP + FP}.
\end{equation}

The Recall metric, conversely, is equivalent to the true positive rate, representing the ratio of accurately classified deforestation pixels to the total number of original deforestation pixels.

\begin{equation}
    \textrm{Recall} = \frac{TP}{TP + FN}.
\end{equation}

With Recall and Precision, the F1-Score is calculated as follows:

\begin{equation}
    \textrm{F1-Score} = \frac{2 \cdot \textrm{Precision} \cdot \textrm{Recall}}{\textrm{Precision} + \textrm{Recall}}.
\end{equation}

\subsection{Computational Resources}

The preliminary experiments were conducted on the following system configuration:

\begin{itemize}
    \item Processor: Thirty-Two-Core AMD Ryzen™ Threadripper™ PRO 5975WX Processor - 3.60GHz 128MB L3 Cache (280W)
    \item Memory: 8 x 64GB PC4-25600 3200MHz DDR4 ECC RDIMM (512GB total)
    \item GPU Accelerators: 3 x NVIDIA® RTX A6000 - 48GB GDDR6 - PCIe 4.0 x16
    \item Operating System: Ubuntu 22.04.2 LTS
\end{itemize}

The use of all GPUs available on the machine for training was enabled by the Mirrored Strategy (\textbf{tf.distribute.MirroredStrategy}\footnote{Available on https://www.tensorflow.org/api$\_$docs/python/tf/distribute/MirroredStrategy}) function of the TensorFlow deep learning framework. This data parallelism approach is intended to accelerate the training process by allowing a deep learning model to be replicated across multiple GPUs, where each GPU retains a full copy of the model. During training, each replica processes a portion of the training data, and then gradient updates are synchronized between GPUs to update the global model. According to \cite{pang2020deep}, using this function further accelerates training and allows for larger models by leveraging memory from multiple GPUs.

\section{Experimental Results}\label{sec:results}

\begin{figure}[!ht]
\centering
\includegraphics[width=0.6\textwidth]{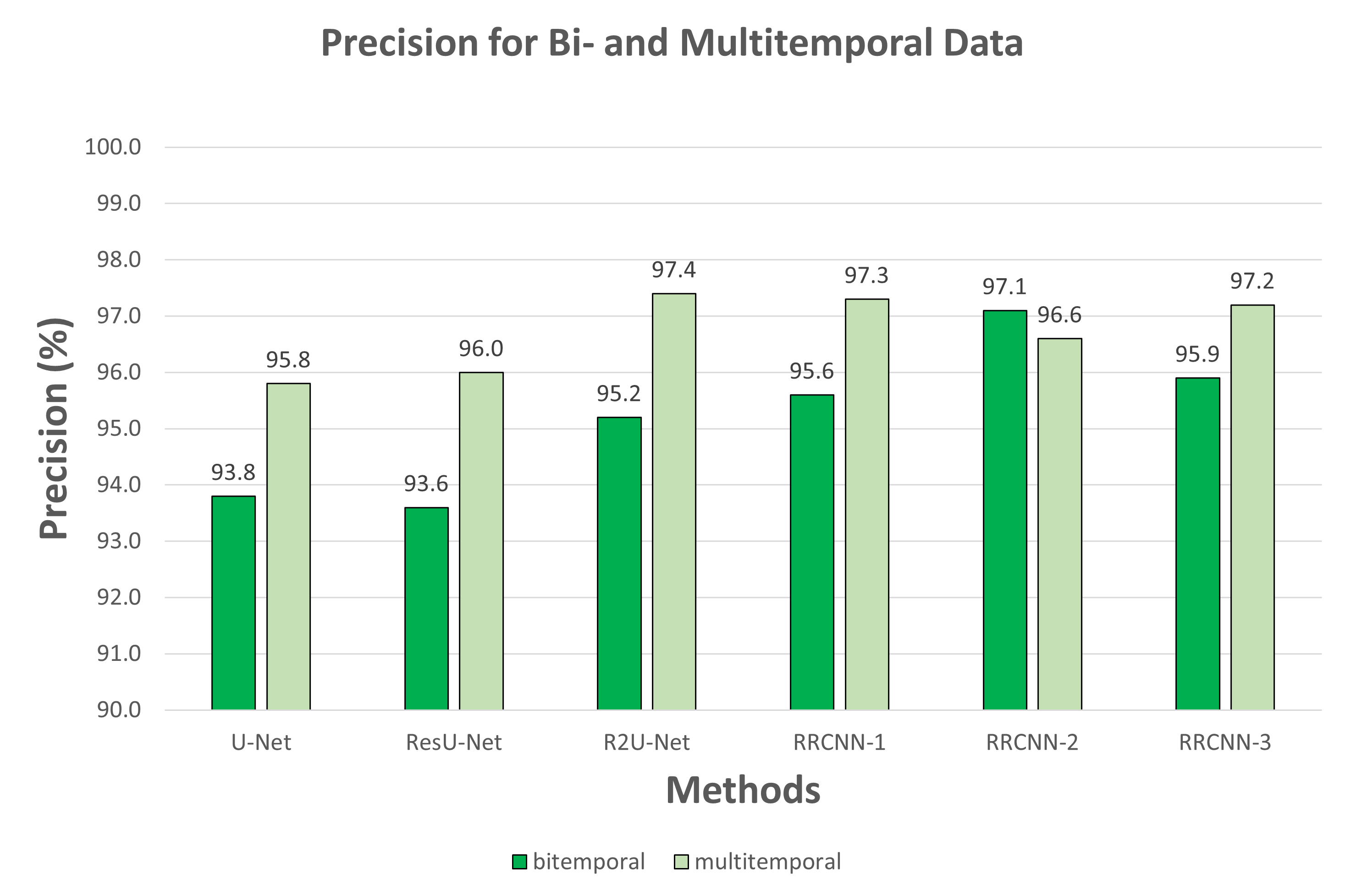}
\caption{Comparison of methods in terms of Precision for bitemporal and multitemporal data}
\label{fig:Precision}
\end{figure}

\begin{figure}[!ht]
\centering
\includegraphics[width=0.6\textwidth]{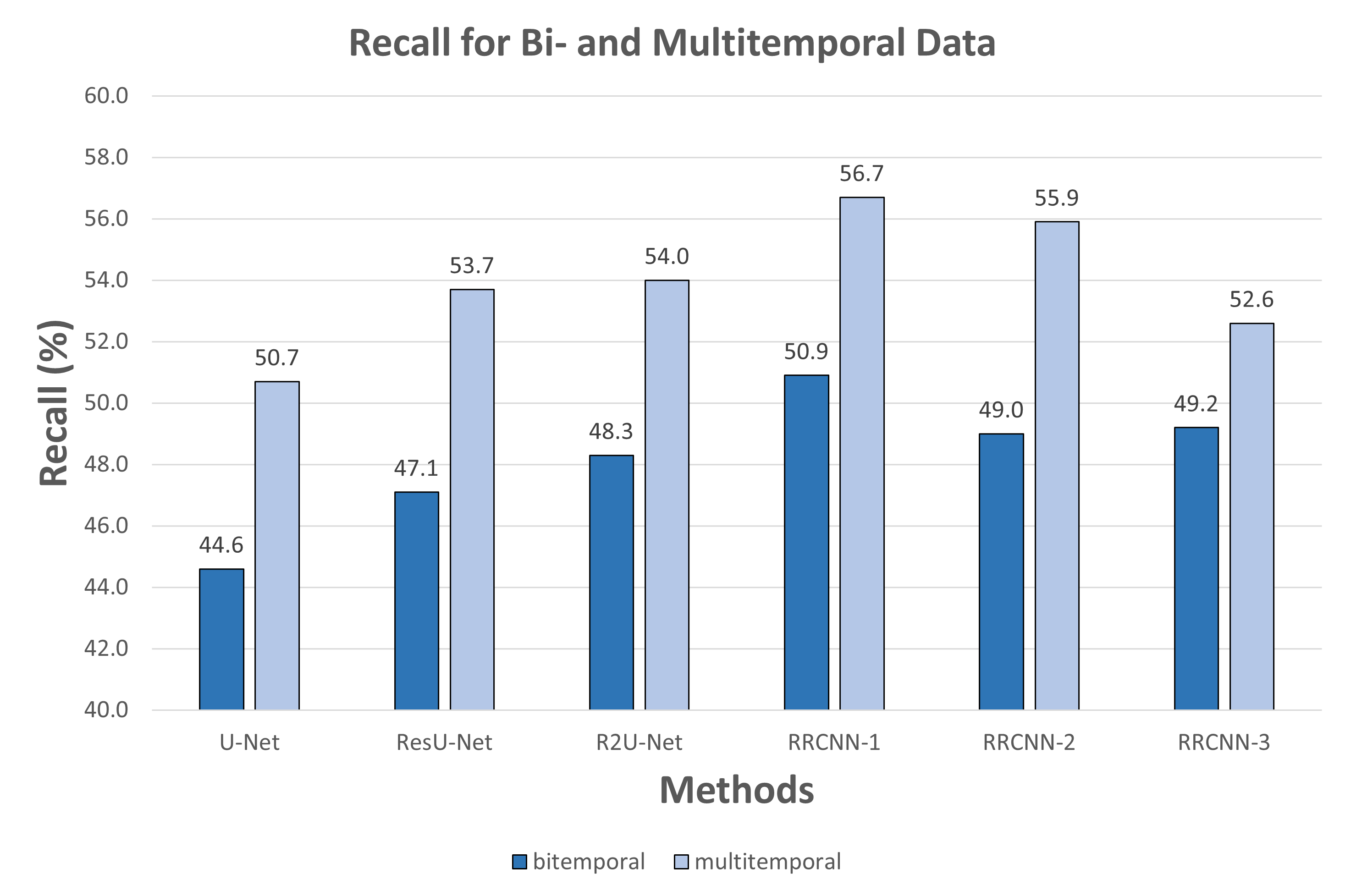}
\caption{Comparison of methods in terms of Recall for bitemporal and multitemporal data}
\label{fig:Recall}
\end{figure}

\begin{figure}[!ht]
\centering
\includegraphics[width=0.6\textwidth]{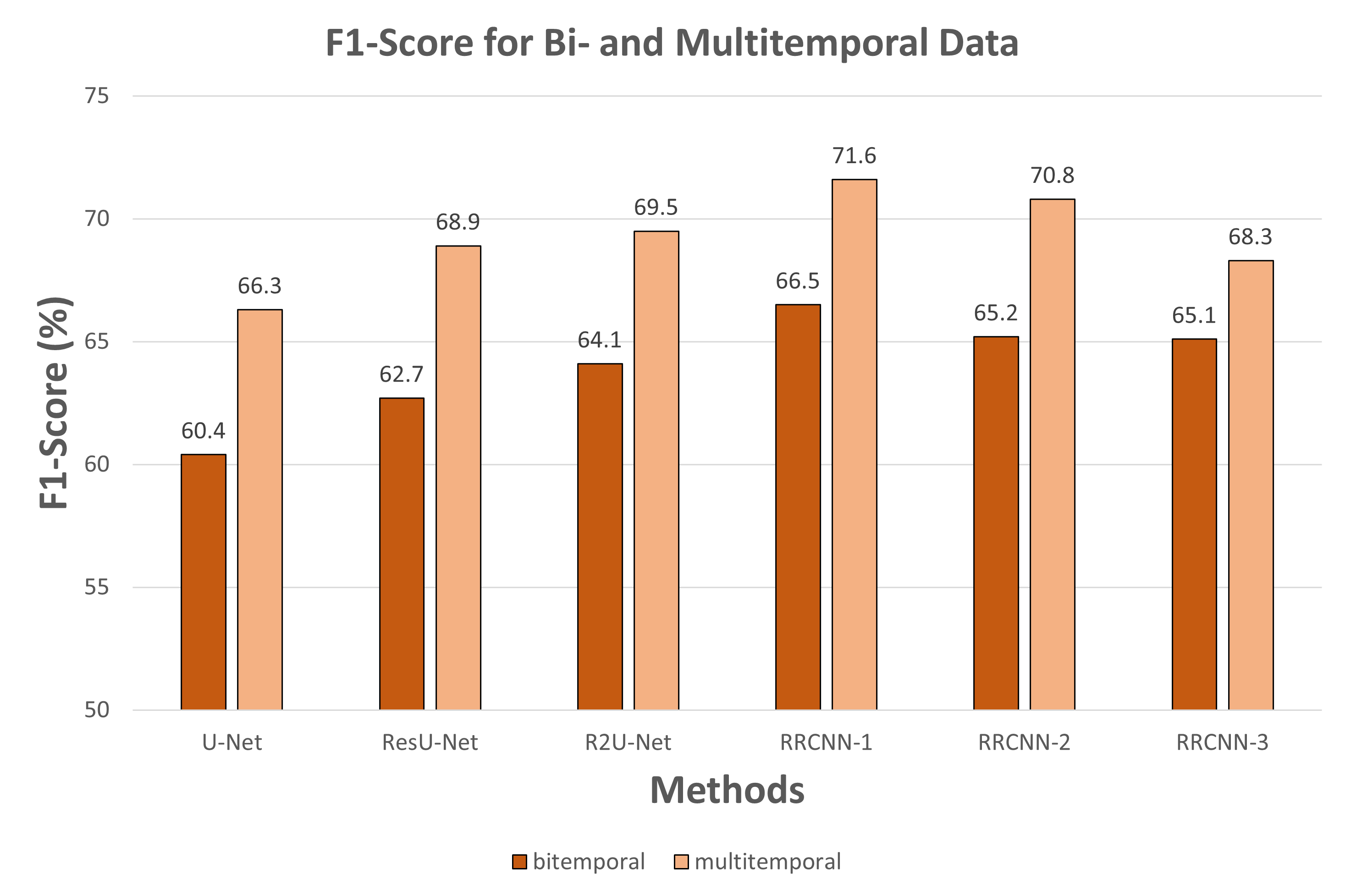}
\caption{Comparison of methods in terms of F1-Score for bitemporal and multitemporal data}
\label{fig:F1Score}
\end{figure}

In this section, we delve into the findings derived from our experimental analysis to evaluate the performance of the deforestation detection architectures outlined in Section \ref{sec:proposal}. Figures \ref{fig:Precision} through \ref{fig:F1Score} showcase the performance metrics of the methods introduced in this study, alongside those referenced in Section \ref{sec:theory}, which serve as our baseline for comparison. These figures provide a comprehensive assessment of accuracy, encompassing Precision, Recall, and F1-score, derived from experiments on both bitemporal and multitemporal input data. The reported scores have been derived using K-fold cross-validation with a value of $K=6$, following the methodology detailed in Section \ref{sec:preparation}.

The first finding that emerges from the analysis of the figures is that all performance metrics derived from multi-temporal data consistently outperformed those obtained from bi-temporal data. Notably, the only exception was the Precision for RRCNN-2, which declined by a small amount for the multitemporal data. This observation corroborates the hypothesis that signs of change in SAR images get less apparent with time. Consequently, using a sequence rather than a mere pair of SAR images improves the chance of capturing changes that occur during the target observation interval.

The second conclusion from our experiments is that recurrent variants, namely the R2U-Net and the three RRCNN variants, consistently outperformed their strictly convolutional counterparts, namely U-Net and ResU-Net. This trend was apparent in all three metrics we examined.

As for the three proposed variants, the RRCNN-1 consistently outperformed the other two, with RRCNN-2 holding a slight edge over RRCNN-3. Remarkably, RRCNN-1 showcased the best results among all the architectures we examined, surpassing the top-performing baseline, the R2U-Net, by 2\% in F1-Score.

\begin{figure}[!ht]
\centering
\includegraphics[width=0.6\textwidth]{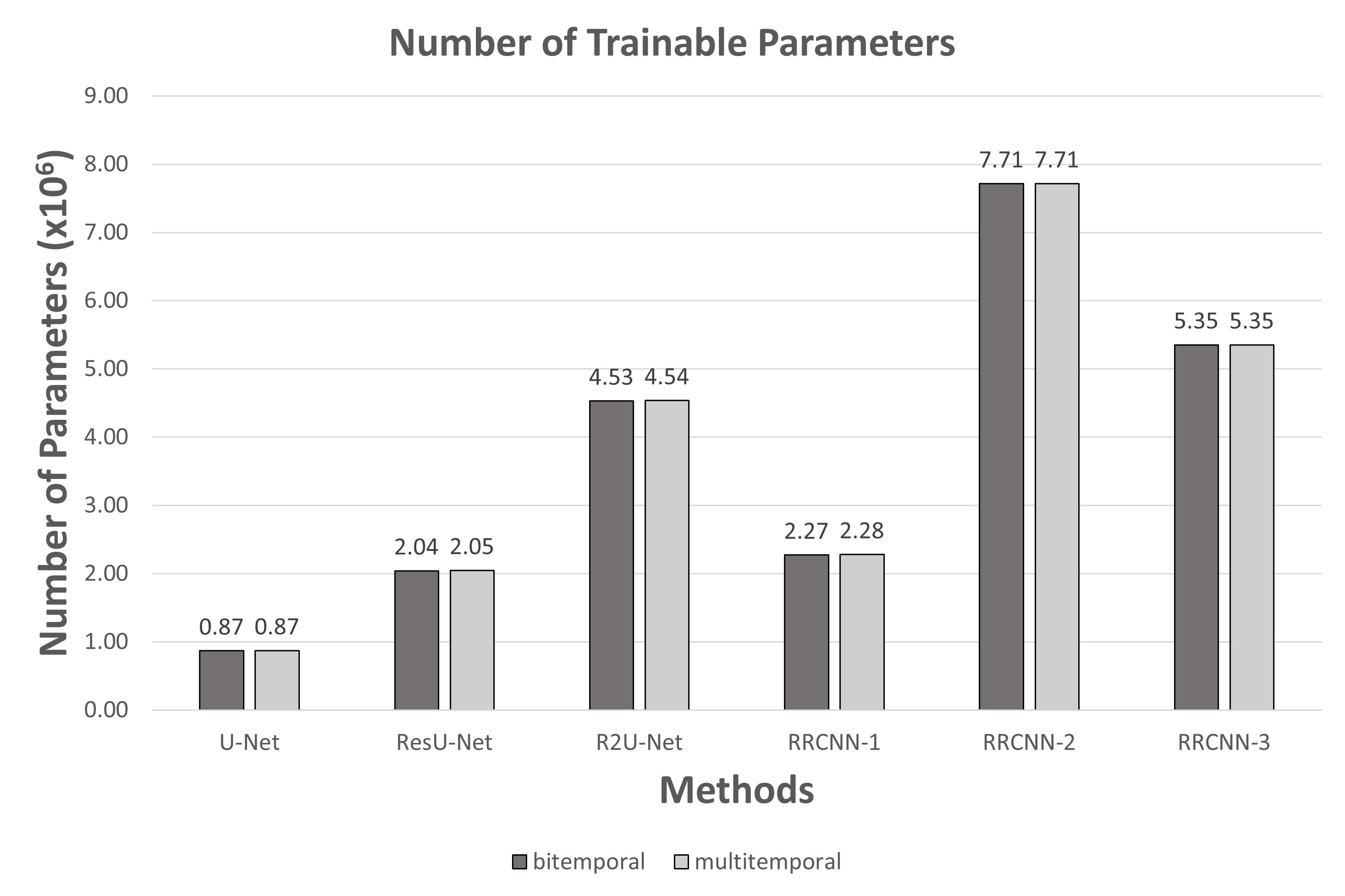}
\caption{Number of Parameters of each model for the bitemporal and multitemporal data}
\label{fig:Parameters}
\end{figure}

Another aspect deserving of examination is the computational efficiency. Figure  \ref{fig:Parameters} provides insight into the count of trainable parameters associated with each scrutinized configuration. 

By looking at each bar group in Figure \ref{fig:Parameters}, one observes that adopting a sequence of images instead of a mere image pair had a marginal impact on the parameter count for each model. It is also observed that the RRCNN-2 and RRCNN-3 were the variants with the largest parameter count, followed closely by R2U-Net. 

Interestingly RRCNN-1 stands out by carrying approximately half the number of parameters when contrasted with R2U-Net, bringing it close to the parameter count of ResU-Net. This implies that RRCNN-1 successfully incorporates recurrence into its model framework with minimal alterations to the overall parameter load compared to ResU-Net, a fully convolutional network.

\begin{figure}[!ht]
\centering
\includegraphics[width=0.6\textwidth]{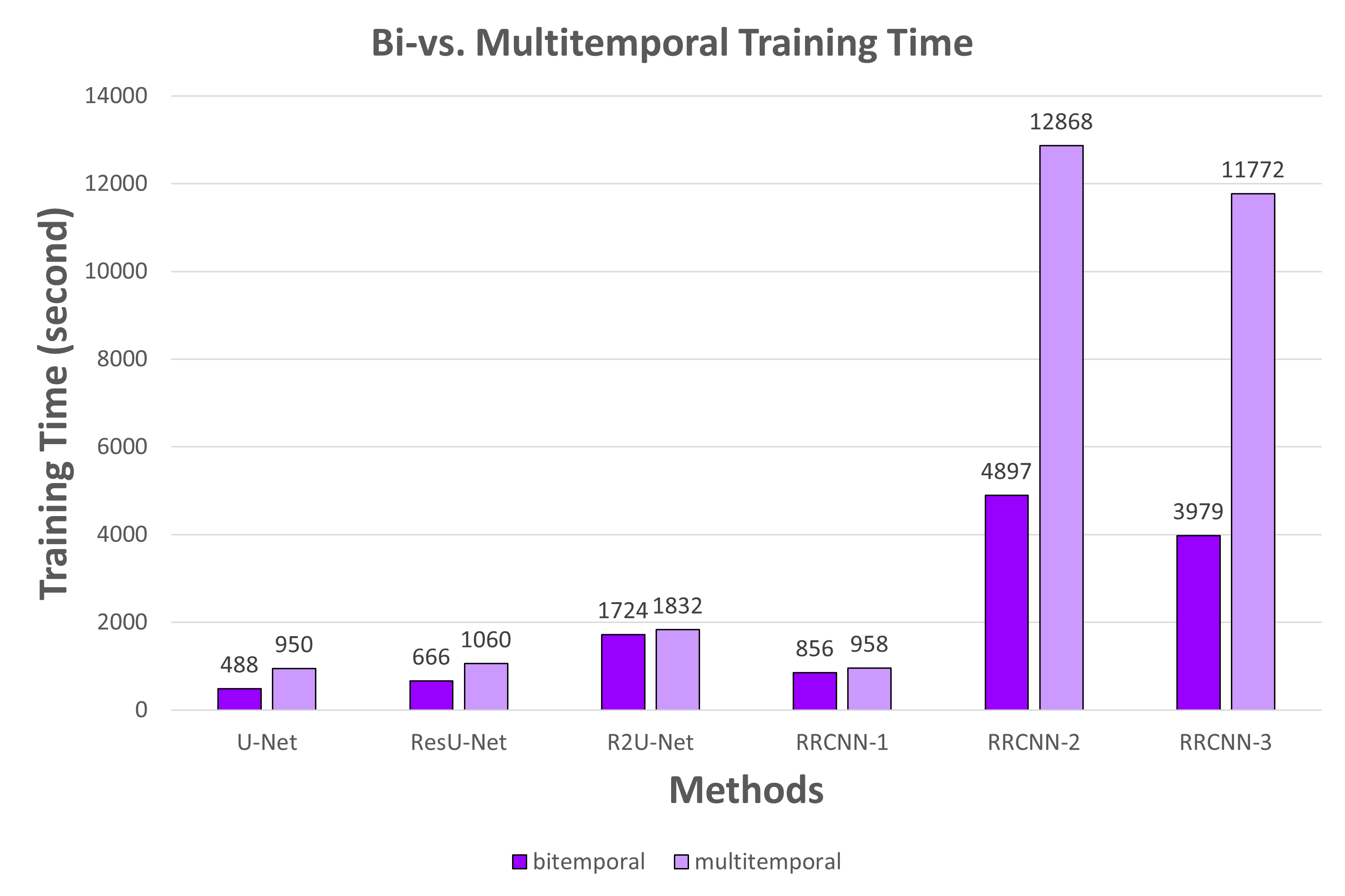}
\caption{Training Time of each model for bitemporal and multitemporal data}
\label{fig:Training}
\end{figure}

\begin{figure}[!ht]
\centering
\includegraphics[width=0.6\textwidth]{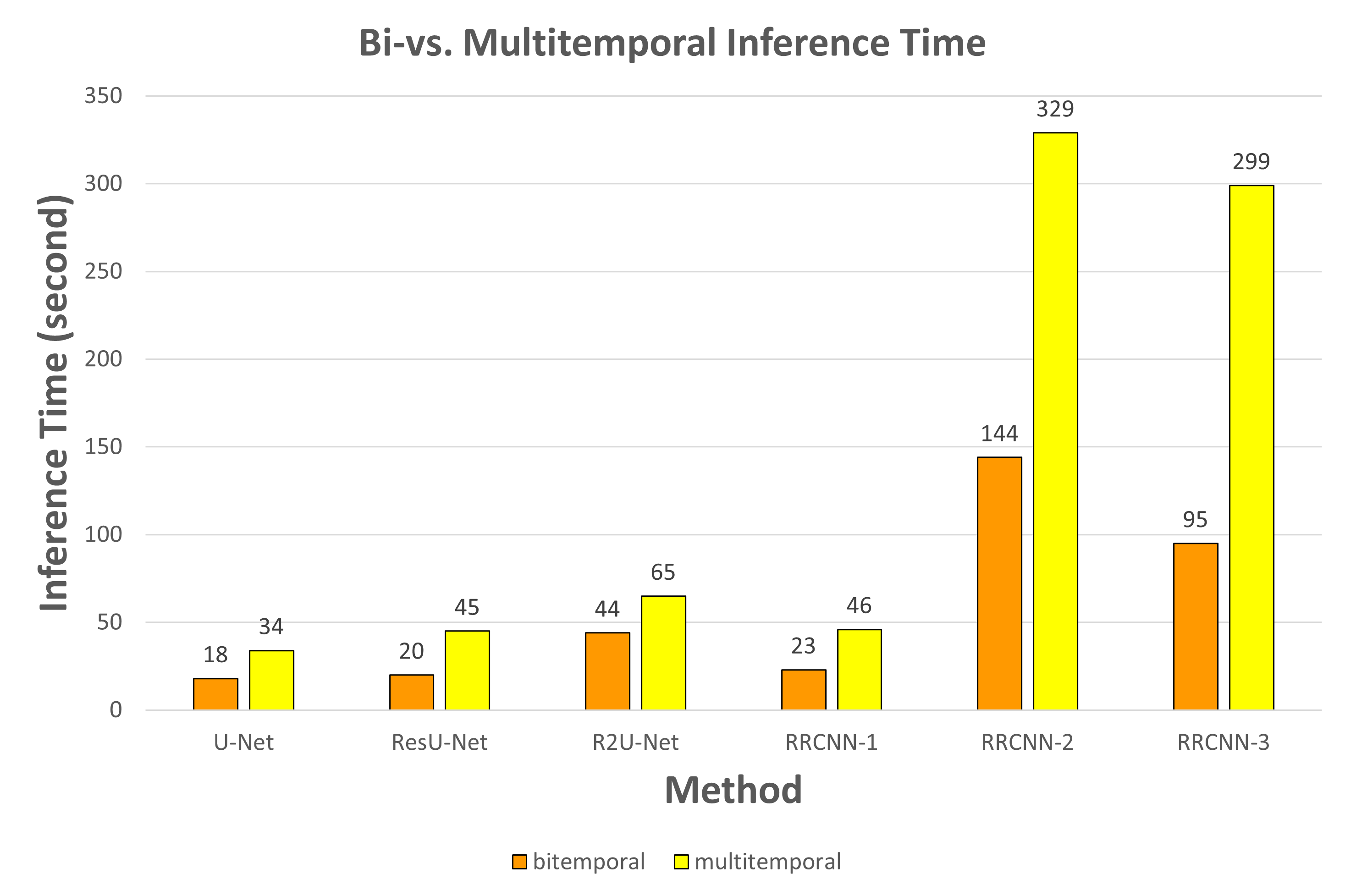}
\caption{Inference Time of each model for bitemporal and multitemporal data}
\label{fig:Inference}
\end{figure}

Figure \ref{fig:Training} and Figure \ref{fig:Inference} show the training and inference times. By and large, as expected, these Figures show a  profile similar to those of the parameter values. 

As observed in Figures \ref{fig:Precision} through \ref{fig:F1Score} together with Figures \ref{fig:Training} - \ref{fig:Inference},  RRCNN-1 very clearly presented the best compromise between accuracy and computational efficiency. Among the variants analyzed, RRCNN-1 presented the best accuracy, and, on the other hand, processing times, particularly the inference time, were close to that of the fastest network, the U-Net.

\begin{figure}
\centering
\includegraphics[width=0.75\textwidth]{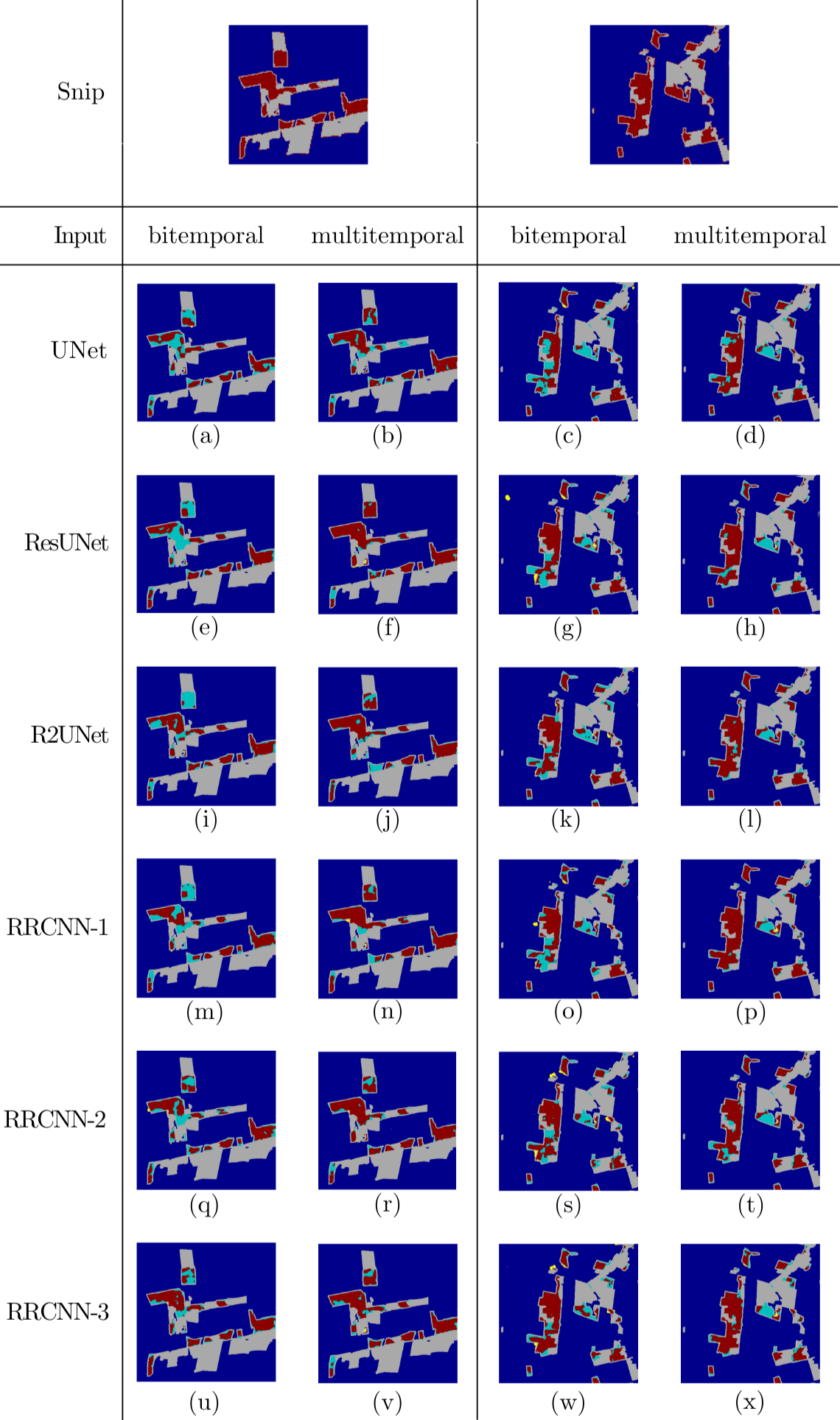}
\caption[Predicted deforestation maps]{Predicted deforestation maps in two snips from the test set. Legend - {\color[HTML]{656565}  past deforestation}; {\color[HTML]{9A0000}  deforestation (true positives)}; {\color[HTML]{00009B} no deforestation (true negatives)}; {\color[HTML]{FFC702} false positives}; {\color[HTML]{2AEBE4} false negatives}.}
\label{fig:changemaps}
\end{figure}

To evaluate the deforestation maps generated in the experiments, Figure \ref{fig:changemaps} illustrates the change maps produced in each experiment in snips of the target site. Each row refers to a different architecture, and each column shows the results for a different snip obtained with the bitemporal and multitemporal inputs.

As observed in Figure \ref{fig:changemaps} the false positive spots are minimal, being noticed only in minor regions, mainly in maps generated with bitemporal input (Figure \ref{fig:changemaps}-(g), (k), (o), (s) and (w)) and even smaller spots in Figure \ref{fig:changemaps}-(p) and (v), that represent maps generated with a multitemporal input.

Regarding false negative spots, their decrease is noticed in the maps generated with multitemporal inputs in relation to bitemporal inputs, confirming the quantitative results. The change map generated by the RRCNN-1 with the multitemporal input was the closest to the ground-truth, as can be seen in Figure \ref{fig:changemaps}(p). The other developed architectures, RRCNN-2 and RRCNN-3, also delivered well defined maps.

\section{Conclusions and Future Work }\label{sec:Concl}

The present research seeks for developing solutions for deforestation monitoring using deep learning aproaches. Until the present stage of this investigation, three change detection architectures relying on recurrent residual learning have been formulated, RRCNN-1, RRCNN-2 and RRCNN-3. These methods were compared with three techniques from the literature, U-Net, ResU-Net and R2U-Net, the latter being a residual recurrent network.

Preliminary experiments were conducted using Sentinel-1 SAR images corresponding to a region of the Brazilian Amazon rainforest. The ground-truth used in this work was collected from the PRODES Project, which was developed by the National Institute for Space Research (INPE). The performance of the techniques was compared using bitemporal data, which are usually emphasized in the literature reports, and also multitemporal data.

RRCNN-1 presented the best performance in most metrics, achieving a F1-Score of 66,5\% with the bitemporal input and 71,6\% with the multitemporal input. RRCNN-2 had the best Precision (97,1\%), the second best F1-Score (65,2\%) and the third best Recall (49,0\%) with the bitemporal input and the second best Recall and F1-Score with the multitemporal input, 55,9\% and 70,8\%, respectively. RRCNN-3 achieved the second best Recall (49,2\%) and Precision (95,9\%) in the bitemporal case and the third best Precision (97,2\%) with the multitemporal input. 

Based on the assessed metrics and the change maps generated through the tested networks, it became evident that the incorporation of an  extended sequence of images significantly enhanced the deforestation detection performance, highlighting the potential benefits of incorporating a comprehensive longer temporal context in the analysis.  RRCNN-1, particularly, delivered significant improved results with a multitemporal input compared to the bitemporal case, while incurring a training time increase of nearly 2 minutes.

The RRCNN-2 and RRCNN-3 networks are designed with ConvLSTM layers in their architectures, which led to longer training and inference times. This is attributed to the inherently high computational complexity associated with LSTM operations, resulting in more demanding computational resources during the learning process.

The next steps for this research include tests with other datasets commonly employed in deforestation detection surveys.


\section*{Acknowledgments}
\label{acknowledgments}
The authors would like to thank the financial support provided by CNPq, CAPES and FAPERJ.



\if true
This section will discuss the  results obtained in this study with the deforestation detection architectures proposed in Section \ref{sec:proposal} for the study area presented in Section \ref{sec:dataset}. Comparisons with counterpart neural network models discussed in Chapter \ref{sec:theory} are also presented. To compute the scores it is applied the K-fold cross-validation with $K=6$, following the approach of Section \ref{sec:preparation}. The achieved metrics (Section \ref{sec:metrics}), training and inference time and the  predicted deforestation maps obtained in each experiment are reported. 

In this line, Table \ref{tab:performance_bi} shows the performance of the proposed RRCNN-1, RRCNN-2 and RRCNN-3 neural networks in terms of Precision, Recall and F1-Score with a bitemporal input. Besides, the corresponding scores for the baselines U-Net, ResU-Net and R2U-Net are also presented in Table \ref{tab:performance_bi}.

\begin{table}[!ht]
\centering
\caption[Performance evaluation (bitemporal)]{Precision, Recall and F1-Score achieved with a bitemporal input.}
\label{tab:performance_bi}
\setlength\extrarowheight{3pt}
\resizebox{\linewidth}{!}{
\begin{tabular}{lcccccc}
\hline
Method    & U-Net & ResU-Net & R2U-Net & RRCNN-1       & RRCNN-2       & RRCNN-3       \\
\hline
Recall    & 44.6 & 47.1    & 48.3   & \textbf{\textcolor{violet}{50.9}} &  \textbf{\textcolor{red}{49.0}} & \textbf{\textcolor{purple}{49.2}}   \\
Precision & 93,8 & 93,6    & 95.2   & \textbf{\textcolor{red}{95.6}} & \textbf{\textcolor{violet}{97.1}} &  \textbf{\textcolor{purple}{95.9}}    \\
F1-Score  & 60,4 & 62,7    & 64.1   &  \textbf{\textcolor{violet}{66.5}} &  \textbf{\textcolor{purple}{65.2}}   & \textbf{\textcolor{red}{65.1}}\\
\hline
\end{tabular}
}
\caption*{\footnotesize Legend:  \textbf{\textcolor{violet}{best}}, \textbf{\textcolor{purple}{second best}},  \textbf{\textcolor{red}{third best}}}
\end{table}

As observed in the previous table, the architectures proposed in this work demonstrated superior results in the evaluated metrics compared to the considered literature networks. RRCNN-1 achieved the best Recall and F1-Score, followed by RRCNN-3 and RRCNN-2. The latter obtained the best result for the Precision metric, followed by RRCNN-3 e RRCNN-1.

Table \ref{tab:time_bi} shows the number of parameters with the training and inference time for each network. The changing of R2U-Net decoder that resulted in RRCNN-1 showed that the network with fewer parameters had better results than the original one, increasing the Recall in 2.6\%, the Precision in 0.5\%, and the F1-Score in 2.5\%.

The RRCNN-3 model, derived from the proposed RRCNN-2 network with a modification in the bottleneck, also achieved comparable results with fewer parameters, as observed from Tables \ref{tab:performance_bi} and \ref{tab:time_bi}.

\begin{table}[!ht]
\centering
\caption[Training time evaluation (bitemporal)]{Number of Parameters, training and inference time (seconds) using a bitemporal input}
\label{tab:time_bi}
\setlength\extrarowheight{3pt}
\resizebox{\linewidth}{!}{
\begin{tabular}{lcccccc}
\hline
Method         & U-Net     & ResU-Net   & R2U-Net    & RRCNN-1   & RRCNN-2   & RRCNN-3   \\
\hline
Nº Parameters  & 868,483  & 2,041,283 & 4,530,179 & 2,272,259 & 7,713,827 & 5,353,507 \\
Training time  & 488 & 666 & 1724 & 856 & 4897 & 3979 \\
Inference time & 18 & 20 & 44 & 23 & 144 & 95 \\
\hline
\end{tabular}
}
\end{table}

U-Net and ResU-Net had the shortest training and testing time, followed by RRCNN-1. In addition to improving the evaluated metrics, RRCNN-1 used half the training time of R2U-Net. The RRCNN-2 and RRCNN-3 models, which use convolutional LSTM in their architecture, had a longer training and inference time.

\begin{table}[!ht]
\centering
\caption[Performance evaluation (multitemporal)]{Precision, Recall and F1-Score achieved with a multitemporal input.}
\label{tab:performance_multi}
\setlength\extrarowheight{3pt}
\resizebox{\linewidth}{!}{
\begin{tabular}{lcccccc}
\hline
Method    & U-Net & ResU-Net   & R2U-Net        & RRCNN-1       & RRCNN-2      & RRCNN-3       \\
\hline
Recall    & 50,7 & 53,7      & \textbf{\textcolor{red}{54,0}} & \textbf{\textcolor{violet}{56,7}} &  \textbf{\textcolor{violet}{55,9}}    & 52,6 \\
Precision & 95,8 & 96,0      & \textbf{\textcolor{violet}{97,4}} &  \textbf{\textcolor{purple}{97,3}}     & 96,6         & \textbf{\textcolor{red}{97,2}} \\
F1-Score  & 66,3 & 68,9      &  \textbf{\textcolor{red}{69,5}}         & \textbf{\textcolor{violet}{71,6}} &  \textbf{\textcolor{purple}{70,8}}    & 68,3   \\
\hline
\end{tabular}
}
\caption*{\footnotesize Legend: \textbf{\textcolor{violet}{best}},  \textbf{\textcolor{purple}{second best}}, \textbf{\textcolor{red}{third best}}}
\end{table}

The result in the evaluated metrics improved when using the multitemporal input with a larger number of images. This improvement was especially seen in the Recall and, consequently, in the F1-Score. As in the case of the bitemporal input, RRCNN-1 had the best performance in Recall and F1-Score, which increased by almost 6\% to the results shown in Table \ref{tab:performance_bi}. RRCNN-2 had the second-best result in these same metrics, increasing Recall by 6.9\% and F1-Score by 5.6\%. R2U-Net increased its Precision by more than 2\% with multitemporal input and had the best result on this metric, followed by RRCNN-1 and RRCNN-3.

Table \ref{tab:time_multi} shows the number of parameters and training time for each network. 

\begin{table}[!ht]
\centering
\caption[Training time evaluation (multitemporal)]{Number of Parameters, training and inference time (seconds) using a multitemporal input} 
\label{tab:time_multi}
\setlength\extrarowheight{3pt}
\resizebox{\linewidth}{!}{
\begin{tabular}{lcccccc}
\hline
Method         & U-Net    & ResU-Net   & R2U-Net    & RRCNN-1   & RRCNN-2   & RRCNN-3 \\
\hline

Nº Parameters  & 871.363  & 2.047.043 & 4.536.579 & 2.278.659 & 7.713.827 & 5.353.507 \\
Training time  & 950 & 1,060  & 1,832  & 958  & 12,868  & 11,772  \\
Inference time & 34 & 45  & 65  & 46  & 329  & 299 \\
\hline
\end{tabular}
}
\end{table}

As expected, training and inference time increased when using more images in the input, especially with RRCNN-2 and RRCNN-3. For each network, the incorporation of a multitemporal input resulted in a training time increment of over two hours in comparison with Table \ref{tab:time_bi}. These networks also had a longer training time than the others models in the bitemporal case, as observed in Table \ref{tab:time_bi}. This was also expected because both architectures use the RCLSTM block, and the high computational complexity typically involved in LSTM cause high hardware demand, in the words of \cite{maor2019fpga}.
\fi

\end{document}